\documentclass{article} 
\usepackage{iclr2027_conference_release,times}

\usepackage{amsmath,amsfonts,bm}

\def\eqref#1{equation~\ref{#1}}

\def\1{\bm{1}}

\DeclareMathAlphabet{\mathsfit}{\encodingdefault}{\sfdefault}{m}{sl}
\SetMathAlphabet{\mathsfit}{bold}{\encodingdefault}{\sfdefault}{bx}{n}

\usepackage{hyperref,url}
\usepackage{booktabs,multicol,multirow,array}
\usepackage{xspace,xcolor,enumitem}
\usepackage{etoolbox,tikz,pgfplots,pgf-pie}
\usepackage[most]{tcolorbox}
\usetikzlibrary{arrows.meta,calc,matrix,positioning}
\pgfplotsset{compat=1.18}

\newcommand{\ourbenchmark}{\textsc{FormalTCS}\xspace}
\newcolumntype{L}[1]{>{\raggedright\arraybackslash}p{#1}}

\definecolor{softblue}{RGB}{100,150,200}
\newtcolorbox[auto counter, number within=section]{prompt}[2][]{%
  colback=white, 
  colframe=softblue!150, 
  width=\textwidth, 
  arc=3mm, 
  boxrule=0.8mm, 
  title=\normalsize #2, 
  breakable=false, 
  fonttitle=\small, 
  fontupper=\footnotesize, 
  #1 
}

\makeatletter
\patchcmd{\pgfpie@legend}
  {yshift={(\the\pgfpie@sliceLength*0.5+1)*0.5cm}}
  {yshift={(\the\pgfpie@sliceLength*0.5+1)*0.5cm-0.23cm}}{}{}
\makeatother

\title{\ourbenchmark: Benchmarking End-to-End\\Frontier Formal Theoretical Computer\\Science Research of Large Language Models}

\author{
    \begin{tabular}[t]{l}
        {\bfseries Dingzirui Wang \quad Xuanliang Zhang \quad Keyan Xu \quad Qingfu Zhu \quad Wanxiang Che}\\[2pt]
        \normalfont Harbin Institute of Technology\\
        \normalfont\texttt{\{dzrwang,xuanliangzhang,kyxu,qfzhu,car\}@ir.hit.edu.cn}
    \end{tabular}
}

\begin{document}
    \maketitle
    
    \begin{abstract}
        Large language models (LLMs) have shown growing capabilities for automated theoretical computer science (TCS) research, while existing benchmarks remain far from realistic research settings. 
        We introduce \ourbenchmark, a benchmark for evaluating LLMs on frontier, end-to-end TCS research. \ourbenchmark contains $143$ instances drawn from papers accepted to top-tier TCS conferences in 2025-2026.
        For each instance, we annotate the theorem statements and proof processes in both natural and formal languages to evaluate the main bottlenecks of existing LLMs in TCS.
        Evaluations of leading LLMs reveal that current models remain far from reliably completing the full TCS research pipeline. 
        In particular, we reveal that autoformalization is the sharpest bottleneck of TCS auto research, where the best model achieves only $11.2$ for translating natural-language claims into formal theorem statements, compared with $28.7$ Pass@8 when proving human-provided formal statements. 
        Building on \ourbenchmark, we further develop an automated TCS research framework that generates, formalizes, filters, and proves new claims. 
        Of $64$ generated claims, only $6$ ultimately pass expert evaluation and proof verification, indicating that beyond formalization, limited research taste remains another major barrier to autonomous TCS research\footnote{Our code and benchmark are in \url{https://github.com/zirui-HIT/FormalTCS}.}.
    \end{abstract}
    
    \section{Introduction}
        Theoretical computer science (TCS) studies the fundamental principles of computation through mathematical methods, including models of computation, algorithms, computational complexity, and the limits of computability \citep{leeuwen-1990-handbook,wigderson-2019-mathematics}.
TCS is an important research topic, as it provides the theoretical foundation for understanding which problems can be computed and how efficiently they can be solved.
Given the fundamental importance of TCS and the increasingly strong autonomous research capabilities demonstrated by large language models (LLMs) \citep{chen2025ai4researchsurveyartificialintelligence}, a growing body of work has begun to investigate how well LLMs can perform TCS-related research tasks.
For example, LCS-Bench~\citep{feng2026theoryscaleautoformalizationlogicscomputer} constructs a benchmark from TCS knowledge extracted from textbooks, while TCS-Bench~\citep{cohenaddad2026tcsbenchbenchmarkingstateoftheartgenerative} evaluates LLMs in generating proofs of TCS theorems from natural-language statements.

However, existing TCS benchmarks still have a gap from real-world TCS research due to the following:
\textit{\textbf{(i) Incomplete research process.}}
Existing TCS benchmarks primarily evaluate isolated capabilities, such as autoformalization or proof generation.
They do not provide an end-to-end evaluation of whether LLMs can conduct TCS research from scratch, making it difficult to identify where current models fail throughout the research process.
\textit{\textbf{(ii) Outdated TCS content.}}
Existing benchmarks are largely constructed from textbook material or theorems already available in libraries such as Mathlib~\citep{moura-etal-2021-lean}.
As a result, they provide limited insight and may also suffer from the fact that related material has appeared in the model's training data.
\textit{\textbf{(iii) Simplified theorem settings.}}
Existing benchmarks typically focus on relatively self-contained theorems with complete and explicit definitions.
In contrast, real TCS papers involve paper-specific definitions and assumptions, as well as multi-layered dependencies among lemmas and theorems.
Existing benchmarks, therefore, fall short of measuring whether LLMs can conduct real-world TCS research.

To bridge these gaps, we introduce \ourbenchmark, a benchmark designed to provide a more realistic evaluation of frontier TCS research.
We employ five human experts to collect and annotate theorems from papers accepted by top conferences in TCS.
Compared with existing benchmarks, \ourbenchmark provides a more faithful evaluation of TCS research capabilities:
\textit{\textbf{(i) End-to-end evaluation.}}
\ourbenchmark decomposes TCS research into five stages and evaluates, step by step, an LLM's ability to transform a natural-language TCS core claim into a corresponding rigorous Lean proof.
This design enables a fine-grained diagnosis of the capabilities and bottlenecks of LLMs throughout the TCS research process.
\textit{\textbf{(ii) Frontier research content.}}
\ourbenchmark is constructed from papers accepted to FOCS, STOC, SODA, and COLT in 2025 and 2026.
We additionally filter the source papers based on whether they are likely to have been previously exposed to the evaluated LLMs, thereby maintaining the timeliness of the benchmark while reducing the risk of data contamination.
\textit{\textbf{(iii) Realistic research theorems.}}
\ourbenchmark evaluates core theorems drawn directly from real TCS papers while preserving their paper-specific definitions, assumptions, and proof dependencies.
It therefore measures the ability to reason about and prove research-level results more accurately.

\begin{table}[t]
    \centering
    \small
    \caption{
        The main findings revealed by \ourbenchmark.
    }
    \begin{tabular}{>{\raggedright\arraybackslash}m{0.75\textwidth}
                >{\raggedright\arraybackslash}m{0.15\textwidth}}
    \toprule
    \textbf{Finding} & \textbf{Evidence} \\
    \midrule
    Current LLMs remain far from TCS research end-to-end
    & \S\ref{subsec:main_result} \\
    \midrule
    Autoformalization is the sharpest bottleneck of automatic TCS research
    & \S\ref{subsec:main_result} \\
    \midrule
    Current LLMs struggle to generate novel and valuable claims of TCS
    & \S\ref{sec:inspire} \\
    \bottomrule
\end{tabular}

    \vspace{-1em}
    \label{tab:findings}
\end{table}

We evaluate a range of leading LLMs on \ourbenchmark, with the findings summarized in Table~\ref{tab:findings}.
Overall, current LLMs still struggle to perform end-to-end TCS research effectively, demonstrating the necessity of \ourbenchmark.
In particular, we find that \textbf{the primary bottleneck lies in translating a natural-language core claim into appropriate formal definitions and theorem statements}, where the current most advanced LLMs can only achieve the performance of $11.2\%$.
This suggests that the mathematical modeling capabilities of current LLMs remain a major limitation for TCS research.
In addition, building on \ourbenchmark, we develop an end-to-end TCS research framework that supports the complete pipeline from proposing a TCS core claim to producing a rigorous Lean proof.
Our experiments show that existing LLMs can produce rigorous proofs for the core claims they propose themselves.
However, human inspection reveals that most of these claims exhibit limited novelty, suggesting that \textbf{the research taste of current models remains underdeveloped}.

Our contributions can be summarized as follows:
\begin{enumerate}[leftmargin=*]
    \item We introduce \ourbenchmark, a benchmark based on real research theorems for evaluating the end-to-end capabilities of LLMs in frontier TCS research.
    \item Our experiments reveal that a key bottleneck for current LLMs is translating natural-language core claims into appropriate formal definitions and theorem statements, indicating that mathematical modeling remains a major weakness of current LLMs.
    \item Building on \ourbenchmark, we develop an end-to-end research framework for TCS and find that the ability of current LLMs to formulate novel and meaningful research claims remains limited.
\end{enumerate}

    \section{Introduction of \ourbenchmark}
        \subsection{Overall Statistics}
    \begin{figure}
        \centering
        \small
        {\small
\begin{tikzpicture}
    \pie[sum=auto,radius=1.8,rotate=100,
        color={
            blue!30!white,
            green!30!white,
            orange!35!white,
            violet!30!white,
            yellow!40!white,
            red!30!white,
            lime!35!white,
            cyan!30!white,
            teal!30!white,
            magenta!30!white,
            brown!30!white,
            purple!25!white,
            gray!35
        }
    ]{
        26/,
        21/,
        10/,
        16/,
        9/,
        9/,
        9/,
        8/,
        8/,
        8/,
        7/,
        6/,
        6/
    }
    \matrix[matrix of nodes,
        nodes={anchor=west,font=\small,inner sep=1.2pt,
            text depth=0pt,text height=1.6ex},
        column sep=4pt,row sep=5pt,
        right=4mm of current bounding box.east,yshift=0.55pt,anchor=west
    ] {
        \tikz\fill[blue!30!white] (0,0) rectangle (0.22,0.22); & Learning Theory & \tikz\fill[cyan!30!white] (0,0) rectangle (0.22,0.22); & Optimization \\
        \tikz\fill[green!30!white] (0,0) rectangle (0.22,0.22); & Statistics \& Testing & \tikz\fill[teal!30!white] (0,0) rectangle (0.22,0.22); & Complexity \& Lower Bounds \\
        \tikz\fill[orange!35!white] (0,0) rectangle (0.22,0.22); & Fairness \& Privacy & \tikz\fill[magenta!30!white] (0,0) rectangle (0.22,0.22); & Algorithms \& Data Struct. \\
        \tikz\fill[violet!30!white] (0,0) rectangle (0.22,0.22); & Online Learning \& Bandits & \tikz\fill[brown!30!white] (0,0) rectangle (0.22,0.22); & Econ. \& Game Theory \\
        \tikz\fill[yellow!40!white] (0,0) rectangle (0.22,0.22); & Coding \& Communication & \tikz\fill[purple!25!white] (0,0) rectangle (0.22,0.22); & Algebra \& CSP \\
        \tikz\fill[red!30!white] (0,0) rectangle (0.22,0.22); & Graphs \& Combinatorics & \tikz\fill[gray!35] (0,0) rectangle (0.22,0.22); & Geometry \& Clustering \\
        \tikz\fill[lime!35!white] (0,0) rectangle (0.22,0.22); & Neural \& Generative Models &  &  \\
    };
\end{tikzpicture}
}
        \vspace{-2em}
        \caption{
            The distribution of research areas covered by \ourbenchmark.
        }
        \vspace{-1.5em}
        \label{fig:category}
    \end{figure}
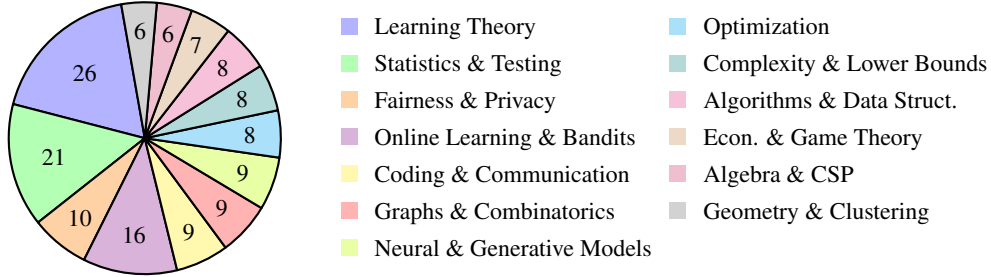

    \ourbenchmark is designed to evaluate the capabilities of LLMs in frontier end-to-end TCS research.
    It consists of $143$ instances, each derived from a distinct research paper, covering $143$ papers in total.
    We use Lean 4.32.2~\citep{moura-etal-2021-lean} together with the corresponding version of Mathlib, which is the latest version available at the time of annotation.
    We ensure the quality of \ourbenchmark along the following dimensions:
    \textit{\textbf{(i) High difficulty.}}
    Across all instances, the expert-validated Lean proofs contain an average of $22.0$ statements, $29.6$ nodes, and $1888.1$ proof lines, indicating that the benchmark involves substantial formalization and proof complexity.
    \textit{\textbf{(ii) High diversity.}}
    As shown in Figure~\ref{fig:category}, \ourbenchmark spans $13$ major research areas in TCS.
    This broad coverage enables the benchmark to evaluate LLM research capabilities across a diverse range of TCS research.

\subsection{Data Format}
    \begin{table}[t]
        \centering
        \small
        \caption{
            Data fields of \ourbenchmark.
        }
        \setlength{\tabcolsep}{6pt}
\renewcommand{\arraystretch}{1.2}
\begin{tabular}{llll}
    \toprule
    \textbf{Category} & \textbf{Name} & \textbf{Type} & \textbf{Meaning} \\
    \midrule
    \multirow{5}{*}{\textbf{Metainfo}}
     & \texttt{id} & string & Data id. \\
     & \texttt{conference} & string & Conference of the accepted paper. \\
     & \texttt{year} & int & Accepted year of the paper. \\
     & \texttt{paper} & string & Name of the paper. \\
     & \texttt{core\_label} & string & LaTeX label in the paper of claim used. \\
    \midrule
    \multirow{3}{*}{\textbf{Natural Language}}
     & \texttt{core\_claim} & string & Core finding of \texttt{core\_label} claim. \\
     & \texttt{nl\_theorem} & string & Full statement of \texttt{core\_label} claim. \\
     & \texttt{nl\_proof} & string & Proof sketch of \texttt{core\_label} claim. \\
    \midrule
    \multirow{2}{*}{\textbf{Formal Language}}
     & \texttt{fl\_theorem} & Lean file & Theorem to be proved of \texttt{core\_label} claim. \\
     & \texttt{fl\_proof} & Lean project & Full Lean-format proof of \texttt{core\_label} claim. \\
    \bottomrule
\end{tabular}
        \vspace{-1em}
        \label{tab:data_format}
    \end{table}

    The data format of \ourbenchmark is summarized in Table~\ref{tab:data_format}.
    Each instance contains information corresponding to the different stages of the end-to-end TCS research pipeline, enabling us to diagnose the capabilities and bottlenecks of current LLMs at each stage of the research process.
    Importantly, every instance is accompanied by a rigorous Lean proof that has been manually verified by experts, which ensures the correctness and reliability of the formalization and, consequently, the overall quality of \ourbenchmark.
    We provide representative cases from \ourbenchmark in Appendix~\ref{app:benchmark_case_study}.

    \section{Annotation of \ourbenchmark}\label{sec:annotation}
        \begin{figure*}
    \centering
    \small
    \includegraphics[width=0.72\linewidth]{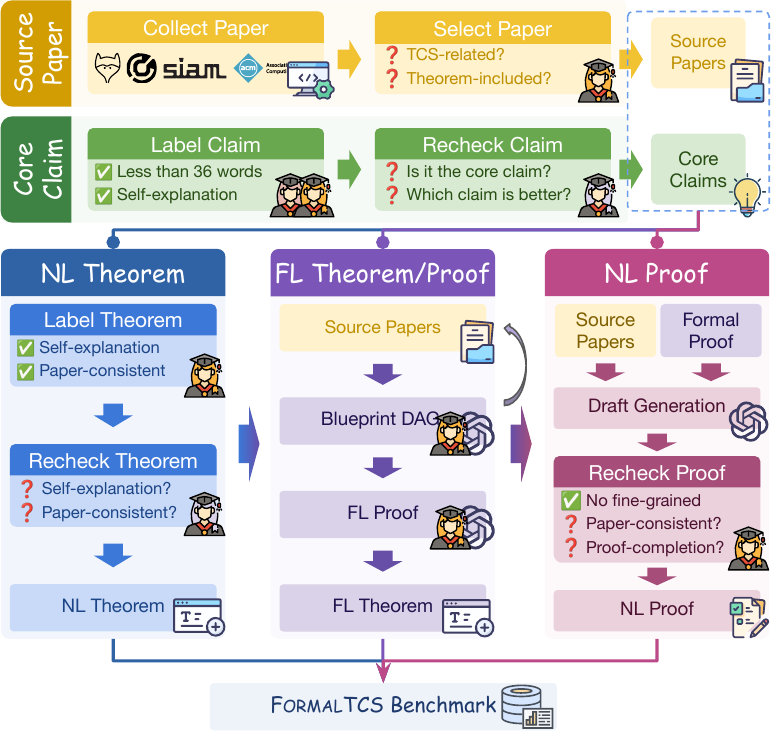}
    \caption{
        The annotation pipeline of \ourbenchmark.
    }
    \vspace{-1.5em}
    \label{fig:annotate_pipeline}
\end{figure*}

This section introduces the annotation pipeline used to construct \ourbenchmark, as illustrated in Figure~\ref{fig:annotate_pipeline}.
Five human experts participate in the annotation process. Each annotator has published multiple papers at top-tier TCS conferences and has substantial research experience in the field.
Given the considerable difficulty of formalizing proofs from the selected papers, we employ \textsc{GPT-5.6-sol} alongside \textsc{Codex} to reduce annotation costs while maintaining data quality.
It should be noted that although the LLM is used as an assistive tool, all final annotations are manually inspected and revised to ensure semantic faithfulness, type correctness, and concise formulations, rather than preserving stylistic artifacts introduced by the LLM.
The prompts used during annotation are provided in Appendix~\ref{app:prompt_annotation}, while Appendix~\ref{app:annotation_info} reports information about the annotators and additional annotation details.
Additional results on human verification pass rates are provided in Appendix~\ref{app:annotate_agree}.

\subsection{Source Paper}
    Our source-paper pool consists of papers accepted to STOC, FOCS, SODA, and COLT in 2025 and 2026.
    This selection is intended to ensure a high level of research quality while reducing the likelihood of benchmark contamination.
    We first use automated scripts to scan all accepted papers and perform an initial filtering step to identify works that appear to study TCS problems.
    Human experts then manually inspect every candidate paper to verify its relevance and determine whether it contains a suitable core result together with a rigorous proof of that result.
    We download the LaTeX source files of these papers from arXiv for subsequent annotation steps.

    To ensure that \ourbenchmark genuinely evaluates TCS research capabilities, we conduct a black-box audit to assess whether the retained papers may have been exposed during model training.
    Specifically, we query \textsc{GPT-5.6-sol} and \textsc{Claude-Opus-5} without retrieval access or paper metadata.
    The inputs consist of partial theorem statements, initial fragments of proofs, and anonymized descriptions of paper results, and the models are asked to reconstruct the missing content.
    We find that the completion similarity of both models is below $9.6\%$, suggesting a relatively low risk of contamination for the selected data.
    The detailed black-box audit is discussed in Appendix~\ref{app:black_box_audit}.

\subsection{Core Claim}
    For each retained paper, human experts produce a concise summary of one of its central theoretical results as overall guidance for the evaluation.
    The summary must contain fewer than $36$ words and should avoid mathematical notation whenever it is not necessary so that the resulting claim is both compact and understandable without additional context.
    When multiple claims from the same paper could reasonably serve as the core claim, annotators select the one that best represents the paper's central contribution.
    This criterion reflects the primary objective of \ourbenchmark, which is to evaluate whether an LLM can recover the relevant theorem and its proof from a given core claim, rather than whether the model can identify which result in a paper is the most important.
    For each paper, two experts independently write candidate core claims.
    A third expert then compares the two candidates and selects the stronger formulation as the final annotation.

\subsection{Natural Language Theorem}
    We next construct a natural-language claim corresponding to the theorem in the source paper associated with the selected core claim.
    Experts first collect all definitions used by the theorem corresponding to the core claim in the source paper. 
    They then rewrite the theorem while preserving its core conclusion to explicitly incorporate these definitions and obtain the natural-language theorem, ensuring that the annotated theorem is self-contained.
    Notably, the theorem corresponding to the core claim may rely on standard domain-specific notions that are not defined in the paper itself (e.g., the multivariate theta transform and the matrix Chernoff bound). We do not supplement such definitions so that \ourbenchmark can more faithfully evaluate LLMs' capabilities for TCS research.
  
\subsection{Formal Language Theorem and Proof}
    In this step, we construct the formal theorem to be proved and its corresponding Lean proof for each instance, providing a reliable basis for evaluating the TCS proof capabilities of LLMs.
    Following previous works \citep{khrulev2026blueprintrepairtypedlocaledits,zhang2026leanmarathonreliableaicomathematicians}, we first use the LLM, together with the source-paper content, to generate a proof-blueprint DAG for the selected core claim.
    Human experts then verify that the resulting blueprint faithfully follows the proof structure of the original paper and that the statement associated with each node is consistent with its corresponding result in the source paper.
    We subsequently invoke the LLM to prove the nodes in the DAG one by one according to their dependency order based on the original paper.
    After each node is proved, a human expert checks that the Lean proof is rigorous and faithful to the corresponding argument in the original paper and verifies that no proof-bypassing constructs such as \texttt{sorry} or additional axioms are used.
    Once the complete Lean proof has been constructed, another human expert performs an independent end-to-end review and fixes any remaining issues.
    The resulting verified artifact is used as the final formal-language proof.
    Finally, we extract from this proof the formal theorem to be proved, together with the definition of closure required to state it, and replace the proof body with \texttt{sorry}.
    The resulting artifact constitutes the formal-language theorem provided to the model as the proof-generation task.

\subsection{Natural Language Proof}
    Our preliminary experiments indicate that current LLMs often struggle to generate rigorous proofs directly from formal-language theorem statements.
    We therefore additionally annotate a natural-language proof sketch, allowing us to diagnose model limitations at a finer level of granularity.
    Each sketch summarizes the main proof ideas used in the source paper, enabling us to evaluate whether an LLM can identify an appropriate high-level proof strategy before carrying out the formal derivation.
    To obtain an initial draft, we provide the LLM with both the verified formal proof and the original paper and ask it to generate a natural-language proof sketch.
    A human expert then checks whether the generated sketch is faithful to the proof strategy in the source paper and whether its reasoning is sufficiently complete.
    Finally, unnecessary low-level derivations and details are removed so that the resulting sketch remains concise while faithfully capturing the overall argument.

    \section{Experiment}\label{sec:experiment}
        In this section, we evaluate a diverse set of mainstream LLMs on \ourbenchmark to investigate the extent to which current models are capable of conducting frontier TCS research.
Rather than treating aggregate benchmark performance as the sole measure of success, we focus on the capabilities required at each stage of the research pipeline and the corresponding failure modes revealed by \ourbenchmark.
This stage-wise evaluation provides a more fine-grained view of where current LLMs perform well in the TCS research workflow and where substantial challenges remain.
All prompts used in our experiments are provided in Appendix~\ref{app:prompt_evaluation}.

\subsection{Experiment Setup}
    \paragraph{Models}
        Given the substantial difficulty of theoretical reasoning tasks, we evaluate three representative families of mainstream LLMs, including \textsc{GPT}~\citep{openai2026gpt56solmodel}, \textsc{Claude}~\citep{anthropic2026opus5systemcard}, and \textsc{DeepSeek}~\citep{deepseekai2026deepseekv4}, using their corresponding harnesses.
        The selected models cover a range of model scales across these families, enabling us to examine both cross-family differences and the relationship between model scale and TCS research performance. 
        This allows for a more comprehensive characterization of the capabilities and limitations of current LLMs on TCS research.
        Detailed model snapshots and harness versions are provided in Appendix~\ref{app:llm_harness_version}.

    \paragraph{Task}
        \begin{table}[t]
            \centering
            \small
            \caption{
                Tasks in \ourbenchmark.
            }
            \begin{tabular}{lll}
    \toprule
    \textbf{Task} & \textbf{Input} & \textbf{Output} \\
    \midrule
    Theorem Elicitation (CC2NT) & Core Claim & NL Theorem \\
    Autoformalization (NT2FT) & NL Theorem & FL Theorem \\
    Proof Elicitation (T2NP) & NL Theorem, FL Theorem & NL Proof \\
    Theorem Proving (FT2FP) & FL Theorem & FL Proof \\
    \bottomrule
\end{tabular}

            \label{tab:task_info}
            \vspace{-1em}
        \end{table}

        \ourbenchmark consists of four tasks that jointly cover the full process of TCS research.
        Table~\ref{tab:task_info} provides detailed definitions of the four tasks.
        To evaluate each stage independently, we provide human-annotated inputs for every task rather than using predictions from the preceding stage as inputs.
        This design prevents errors introduced early in the pipeline from propagating to subsequent tasks, thereby allowing us to identify the sources of performance degradation more precisely.
        The results reported in Table~\ref{tab:main_experiment} further motivate this stage-wise evaluation, as current LLMs remain unable to reliably complete the entire TCS research pipeline in an end-to-end manner.

    \paragraph{Metrics}
        We adopt task-specific evaluation metrics to measure different aspects of capabilities:
        \begin{itemize}[leftmargin=*,nosep]
            \item \textbf{LLM-Rubric~\citep{ma2026reliable} (CC2NT, T2NP)}:
                We adopt an LLM-based rubric to evaluate the semantic consistency.
                Each response receives four scores normalized to the $[0,100]$ range, corresponding to logical validity ($s_{\mathrm{logic}}$), completeness ($s_{\mathrm{complete}}$), correctness ($s_{\mathrm{correct}}$), and clarity ($s_{\mathrm{clear}}$).
                The rubric is calculated as $\mathrm{Score}=0.4s_{\mathrm{logic}}+0.3s_{\mathrm{complete}}+0.2s_{\mathrm{correct}}+0.1s_{\mathrm{clear}}$.
                To reduce evaluation bias, we use \textsc{Qwen3.8-Max} with \textsc{QoderCLI}~\citep{qwen38} as the rubric evaluator, which is distinct from all experimental LLMs.
                We additionally examine the rubric agreement between LLM-based and human evaluations in Appendix~\ref{app:rubric_agree}.

            \item \textbf{BEq+~\citep{poiroux-etal-2025-reliable} (NT2FT)}:
                For the autoformalization task, we adopt BEq+, which determines whether a Lean statement is equivalent to the reference statement through bidirectional proving.
                Given a reference theorem $t_r$ and a generated candidate theorem $t_c$, the metric attempts to prove both $t_r \Rightarrow t_c$ and $t_c \Rightarrow t_r$ in Lean.
                Unlike evaluation methods based on LLM judges, this procedure relies on deterministic symbolic proof search.
                A candidate theorem is considered equivalent to the reference theorem only if proofs in both directions are successfully constructed.

            \item \textbf{Pass@$k$~\citep{dong2024formal} (FT2FP)}:
                For the theorem-proving task, we use Pass@$k$, which measures the proportion of instances for which at least one of the $k$ sampled proofs is accepted by the Lean compiler.
                The metric therefore reflects the probability that the model produces at least one syntactically valid and formally verified proof across multiple generation attempts.
                To prevent bypassing proof obligations, our automated verification environment enables \texttt{set\_option warningAsError true} and uses a custom linter or \texttt{\#print axioms} with an explicit axiom whitelist, thereby ruling out \texttt{sorry}, custom \texttt{axiom} declarations, and similar shortcuts.
        \end{itemize}

    \paragraph{Generation Parameters}
        Following the experimental settings of prior work on formal reasoning~\citep{ren2025deepseekproverv2advancingformalmathematical,lin2025goedelprover}, we generate $8$ candidate outputs for each instance in the NT2FT and FT2FP tasks to balance evaluation cost and reliability while using a single generation for CC2NT and T2NP.
        We adopt different generation settings because Lean outputs admit reliable automatic verification, allowing us to sample multiple formal candidates and evaluate them objectively using symbolic verification.
        In contrast, natural-language responses lack an equally reliable automatic verifier, making single-sample evaluation more appropriate for these tasks.
        For tasks requiring multiple generations, we use a temperature of $0.6$ and set top\_p to $0.9$.
        For single-generation experiments, we use deterministic decoding with a temperature of $0.0$ and top\_p of $1.0$.

\subsection{Experimental Results}\label{subsec:main_result}
    \begin{table}[t]
        \centering
        \small
        \caption{
            Performance of mainstream LLMs with their corresponding harnesses on \ourbenchmark.
            The best performance on each task is marked in \textbf{bold}.
        }
        \begin{tabular}{lll|cccc}
    \toprule
    \textbf{Model} & \textbf{Harness} & \textbf{Scale} &
    \textbf{CC2NT} & \textbf{NT2FT} & \textbf{T2NP} & \textbf{FT2FP} \\
    \midrule
    \multirow{3}{*}{\textsc{GPT-5.6}} & \multirow{3}{*}{\textsc{Codex}} & \textsc{luna} & $56.4$ & $2.8$ & $61.2$ & $14.0$ \\
    &  & \textsc{terra} & $60.7$ & $5.6$ & $64.0$ & $18.2$ \\
    &  & \textsc{sol} & $\mathbf{67.4}$ & $10.5$ & $67.9$ & $26.6$ \\
    \midrule
    \multirow{3}{*}{\textsc{Claude}} & \multirow{3}{*}{\textsc{Claude Code}} & \textsc{Haiku-4.5} & $48.7$ & $2.1$ & $55.3$ & $7.7$ \\
    &  & \textsc{Sonnet-5} & $63.0$ & $9.1$ & $65.7$ & $23.8$ \\
    &  & \textsc{Opus-5} & $66.9$ & $\mathbf{11.2}$ & $\mathbf{68.7}$ & $\mathbf{28.7}$ \\
    \midrule
    \multirow{2}{*}{\textsc{DeepSeek-V4}} & \multirow{2}{*}{\textsc{DeepSeek Harness}} & \textsc{Flash} & $55.6$ & $7.0$ & $61.7$ & $17.5$ \\
    &  & \textsc{Pro} & $58.8$ & $8.4$ & $63.8$ & $21.0$ \\
    \bottomrule
\end{tabular}

        \vspace{-1em}
        \label{tab:main_experiment}
    \end{table}

    Table~\ref{tab:main_experiment} reports the evaluation results.
    Overall, \textsc{Claude-Opus-5} achieves the best performance on most tasks, indicating the strongest TCS research capability.
    Beyond the overall comparison, the results reveal several important findings about the capabilities and limitations of current LLMs.

    \paragraph{Finding 1: Current LLMs Struggle with End-to-End TCS Research}
        Our results show that even the strongest current models remain limited when completing the full TCS research pipeline.
        For example, although the best-performing model, \textsc{Claude-Opus-5}, achieves scores of $66.9$ and $68.7$ on natural-language claim understanding and proof-strategy generation, respectively, substantial bottlenecks remain in the formal stages of the pipeline, with its final formal proof generation performance reaching only $28.7$ Pass@8.
        These results suggest that although current LLMs exhibit meaningful theoretical reasoning capabilities, they still struggle to reliably complete the end-to-end TCS research process from a high-level research claim to a machine-verifiable formal proof.

    \paragraph{Finding 2: Formalization Tasks are more Difficult than Natural-Language Tasks}
        The results show that LLMs perform substantially better on natural-language tasks than on formalization tasks.
        For example, \textsc{Claude-Opus-5} achieves $68.7$ on proof elicitation, whereas its performance on the corresponding theorem formalization task is only $11.2$.
        Similarly, \textsc{GPT-5.6-sol} achieves $67.9$ on proof elicitation but only $10.5$ on theorem autoformalization.
        This substantial performance gap suggests that current LLMs can understand and articulate high-level theoretical ideas considerably better than they can translate those ideas into rigorous formal representations.

    \paragraph{Finding 3: Autoformalization is the Primary Bottleneck for LLM-Based TCS Research}
        Across the entire pipeline, theorem autoformalization is the lowest-performing stage, with no evaluated model exceeding $11.2$.
        In contrast, when provided with a human-annotated formal theorem statement, models achieve up to $28.7$ Pass@8 on the subsequent formal proof generation task.
        This result suggests that the primary difficulty for current models is not merely generating Lean proofs.
        Rather, the more fundamental challenge lies in correctly identifying the mathematical objects, assumptions, and logical structure underlying a natural-language claim and translating them into appropriate formal definitions and theorem statements.
        Improving autoformalization capabilities is therefore a key direction toward enabling LLMs to conduct automated end-to-end TCS research.

    \section{End-to-End Automated TCS Research with \ourbenchmark}\label{sec:inspire}
        \begin{figure*}
    \centering
    \small
    \includegraphics[width=\linewidth]{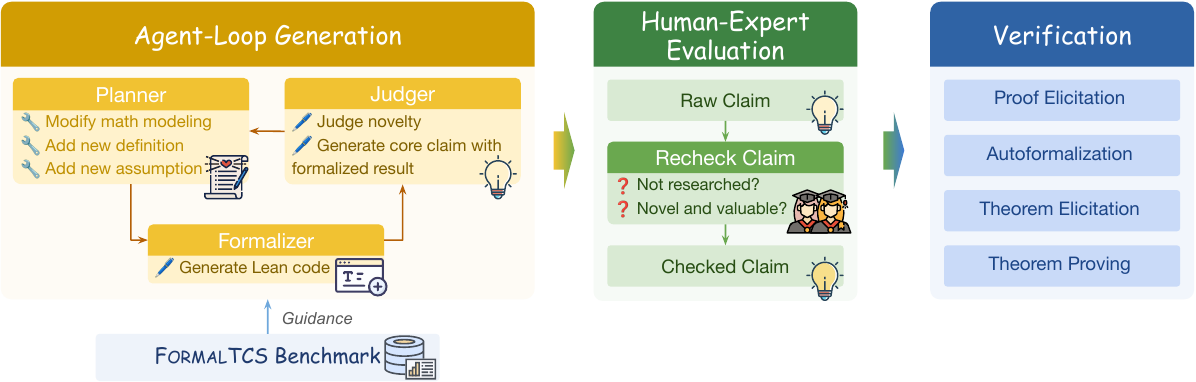}
    \caption{
        Our automatic end-to-end TCS research framework using LLMs based on \ourbenchmark, which consists of three steps:
        \textit{(i) Agent-Loop Generation}: Generating the candidate TCS claims.
        \textit{(ii) Human-Expert Evaluation}: Filtering novel and valuable claims.
        \textit{(iii) Verification}: Generating theorems and proofs based on claims, which is the same as in \S\ref{sec:experiment}.
    }
    \vspace{-1em}
    \label{fig:inspire_pipeline}
\end{figure*}

Although \ourbenchmark primarily focuses on evaluating theoretical reasoning capabilities rather than the quality of newly proposed research ideas, generating meaningful TCS core claims remains an essential component of a fully automated research framework.
Directly evaluating such claims is challenging since determining whether a research idea is genuinely novel is still an open problem and lacks reliable evaluation metrics~\citep{si2025can,si2026the}.
To investigate this capability, we develop a multi-agent framework in this section that enables LLMs to propose candidate claims, translate them into formal statements, and automatically filter them before human evaluation.
In contrast to \S\ref{sec:experiment}, which primarily evaluates the bottlenecks of current LLMs in TCS research, \textbf{this section investigates whether current LLMs, when guided by \ourbenchmark, can autonomously discover valuable TCS claims and produce correct end-to-end proofs for them}.
The overall system is illustrated in Figure~\ref{fig:inspire_pipeline} and the prompts used in this section are provided in Appendix~\ref{app:prompt_generation}.

\subsection{Framework Design}
    \subsubsection{Agent-Loop Generation}
        Our framework is inspired by the iterative nature of real-world TCS research.
        Researchers typically do not commit to a fixed theorem formulation from the outset.
        Instead, they repeatedly revise assumptions, definitions, modeling choices, and proof directions until they identify a result worth pursuing.
        We simulate this iterative process using three agents: a planner, a formalizer, and a judger.
        \paragraph{Planner.}
            Based on the examples of \ourbenchmark and the derivations accumulated so far, the planner proposes a new TCS research objective.
            The objective is not restricted to extending the current line of reasoning.
            The planner may reformulate the objective, introduce auxiliary concepts, strengthen or relax assumptions, or explore alternative analytical directions.
            This flexibility allows the system to search over a diverse space of potential theoretical research directions while keeping each objective plausibly formalizable in Lean 4 with Mathlib.
        \paragraph{Formalizer.}
            The formalizer translates the proposed research objective into a precise Lean statement and repeatedly queries the Lean compiler to identify and correct formalization errors, ensuring that the objective can be strictly formalized.
            For each proposal, we allow at most three rounds of compiler feedback.
            Candidate claims that still fail to compile within this interaction budget are discarded from the subsequent pipeline, and the failure information is returned to the planner.
        \paragraph{Judger.}
            Once a formal statement successfully compiles, the judge translates it back into a concise natural-language claim that summarizes its potential theoretical significance and assesses the value of the proposed result.
            If the claim is judged to lack sufficient novelty, it is discarded, and the corresponding feedback is returned to the planner.
            Claims that pass this filtering stage are added to a candidate pool and subsequently evaluated by human experts.

        All agents operate within a shared workspace using \textsc{GPT-5.6-sol} together with \textsc{Codex}.
        At initialization, the workspace contains only data from \ourbenchmark as the guidance.
        Each agent can freely read or write to the shared workspace while proposing and validating new claims.
        For each claim, every agent maintains exactly one persistent session, allowing the corresponding context to be reused throughout the iterative process.
        To ensure the generation quality and diversity, each run generates only one claim.
        To encourage diversity across generated claims in different runs, at the beginning of each new generation run, we randomly sample $16$ instances from \ourbenchmark, place them in the workspace, and use a temperature of $0.6$ and set top\_p to $0.9$.

    \subsubsection{Human-Expert Evaluation}
        We next conduct a human evaluation of the candidate claims retained after the agent-based generation and filtering process.
        As an initial screening step, human experts remove candidates whose conclusions have already been established in the existing literature.
        The remaining claims are then evaluated according to two criteria: whether they provide a sufficiently novel observation and whether that observation has potential value for TCS research.
        Each candidate claim is independently reviewed by two experts to reduce subjectivity in the evaluation process.
        Since automatically and reliably estimating novelty and research value is itself a difficult research problem, we do not incorporate an automated claim-quality evaluator into the current framework.
        We instead view this capability as an important direction for future work.
        For every claim retained after expert evaluation, we subsequently follow the stage-wise procedure introduced in \S\ref{sec:experiment} to construct its corresponding formal statement and formal proof, ensuring that the resulting claims are rigorously and reliably verified.

\subsection{Experimental Results}
    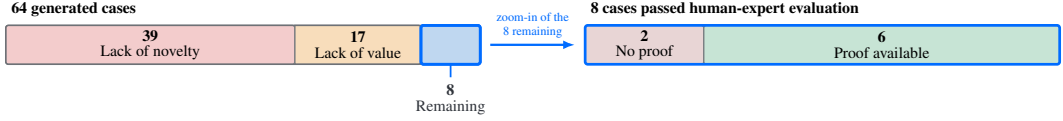
\begin{figure*}
        \centering
        \small
\definecolor{cNovel}{HTML}{E9A3A3}
\definecolor{cValue}{HTML}{F3C78D}
\definecolor{cRemain}{HTML}{A9C9EB}
\definecolor{cZoom}{HTML}{006CFF}
\definecolor{cNoProof}{HTML}{D5B3B3}
\definecolor{cProof}{HTML}{A9D6BF}
\definecolor{cDark}{HTML}{30343B}

\def\NTotal{64}
\def\NNovel{39}
\def\NValue{17}
\def\NRemain{8}
\def\NNoProof{2}
\def\NProof{6}

\resizebox{\linewidth}{!}{%
\begin{tikzpicture}[
    font=\sffamily,
    x=1cm,
    y=1cm
]

\def\H{0.78}
\def\W{9.5}
\def\Gap{2.1}
\def\y{0}

\pgfmathsetmacro{\xB}{\W+\Gap}

\pgfmathsetmacro{\wNovel}{\NNovel/\NTotal*\W}
\pgfmathsetmacro{\wValue}{\NValue/\NTotal*\W}
\pgfmathsetmacro{\wRemain}{\NRemain/\NTotal*\W}

\pgfmathsetmacro{\wNoProof}{\NNoProof/\NRemain*\W}
\pgfmathsetmacro{\wProof}{\NProof/\NRemain*\W}

\node[
    anchor=west,
    font=\bfseries
] at (0,1.10)
{\NTotal\ generated cases};

\node[
    anchor=west,
    font=\bfseries
] at (\xB,1.10)
{\NRemain\ cases passed human-expert evaluation};


\path[
    fill=cNovel!55,
    draw=cDark!65,
    line width=.75pt,
    rounded corners=1.5pt
]
(0,\y)
rectangle
(\wNovel,\y+\H);

\path[
    fill=cValue!60,
    draw=cDark!65,
    line width=.75pt
]
(\wNovel,\y)
rectangle
({\wNovel+\wValue},\y+\H);

\path[
    fill=cRemain!75,
    draw=cZoom,
    line width=1.55pt,
    rounded corners=1.5pt
]
({\wNovel+\wValue},\y)
rectangle
(\W,\y+\H);

\draw[
    cDark!70,
    line width=.85pt,
    rounded corners=1.5pt
]
(0,\y)
rectangle
({\wNovel+\wValue},\y+\H);

\draw[
    cZoom,
    line width=1.55pt,
    rounded corners=1.5pt
]
({\wNovel+\wValue},\y)
rectangle
(\W,\y+\H);


\path[
    fill=cNoProof!55,
    draw=cDark!65,
    line width=.75pt,
    rounded corners=1.5pt
]
(\xB,\y)
rectangle
({\xB+\wNoProof},\y+\H);

\path[
    fill=cProof!65,
    draw=cDark!65,
    line width=.75pt,
    rounded corners=1.5pt
]
({\xB+\wNoProof},\y)
rectangle
({\xB+\W},\y+\H);

\draw[
    cZoom,
    line width=1.55pt,
    rounded corners=1.5pt
]
(\xB,\y)
rectangle
({\xB+\W},\y+\H);


\node[
    align=center,
    font=\small
]
at ({\wNovel/2}, {\y+\H/2})
{
    \textbf{\NNovel}\\[-1pt]
    {\footnotesize Lack of novelty}
};

\node[
    align=center,
    font=\small
]
at ({\wNovel+\wValue/2}, {\y+\H/2})
{
    \textbf{\NValue}\\[-1pt]
    {\footnotesize Lack of value}
};

\node[
    align=center,
    font=\small
]
at ({\xB+\wNoProof/2}, {\y+\H/2})
{
    \textbf{\NNoProof}\\[-1pt]
    {\footnotesize No proof}
};

\node[
    align=center,
    font=\small
]
at ({\xB+\wNoProof+\wProof/2}, {\y+\H/2})
{
    \textbf{\NProof}\\[-1pt]
    {\footnotesize Proof available}
};


\draw[
    cZoom,
    line width=1.0pt
]
({\wNovel+\wValue+\wRemain/2}, {\y})
-- ++(0,-0.26);

\node[
    anchor=north,
    align=center,
    font=\small,
    text=cDark
]
at (
    {\wNovel+\wValue+\wRemain/2},
    {\y-0.28}
)
{
    \textbf{\NRemain}\\[-1pt]
    {\footnotesize Remaining}
};


\draw[
    -{LaTeX[length=2.4mm]},
    line width=1.3pt,
    draw=cZoom
]
(\W+0.20, {\y+\H/2})
--
(\xB-0.20, {\y+\H/2});

\node[
    font=\scriptsize,
    text=cZoom,
    align=center
]
at (
    {\W+\Gap/2},
    {\y+\H/2+0.40}
)
{
    zoom-in of the\\
    \NRemain\ remaining
};

\end{tikzpicture}%
}
        \vspace{-2em}
        \caption{
            The distribution of generated TCS cases using our framework.
        }
        \vspace{-1em}
        \label{fig:inspire_distribution}
    \end{figure*}

    Based on the framework described above, we use the agent loop to synthesize $64$ core claims, of which $6$ remain after human-expert evaluation and proof verification.
    The pass rates at each stage are shown in Figure~\ref{fig:inspire_distribution}.
    We also present two representative cases of the accepted claims in Appendix~\ref{app:synthesis_case_study}.
    These results indicate that, for current LLMs, \textbf{the primary bottleneck in conducting end-to-end TCS research lies in generating claims that are both novel and valuable}, suggesting that their research taste remains limited.
    Therefore, in addition to improving autoformalization capabilities as discussed in \S\ref{subsec:main_result}, advancing the end-to-end TCS research capabilities of LLMs also requires a substantially stronger research taste.
    Interestingly, among the claims that pass human-expert evaluation, the proof success rate is substantially higher than the FT2FP performance reported in Table~\ref{tab:main_experiment}.
    The explanation is that these claims are generated by the LLMs themselves, making it easier for the LLMs to construct proofs for claims that align with their own reasoning trajectories.

    \section{Related Work}
        LLM for TCS refers to the use of LLMs, together with tools such as formal proof assistants, program execution, and search algorithms, to assist with or automate theorem proving, algorithm discovery, and research exploration in theoretical computer science.
Its development can be roughly divided into three stages. Early works, including Autoformalization~\citep{wu-etal-2022-autoformalization}, Draft,~Sketch,~and~Prove~\citep{jiang2023draft}, LeanDojo~\citep{yang-etal-2023-leandojo}, and DeepSeek-Prover~\citep{xin2024advancing,xin2025deepseekproverv}, primarily explored the translation of natural-language mathematics into formal proofs, as well as the use of retrieval, proof-assistant feedback, and search to improve machine-verifiable reasoning capabilities. 
Meanwhile, FunSearch~\citep{Romera-Paredes-etal-2024-mathematical} began combining LLMs with program search to automatically discover new constructions and algorithms for problems in combinatorics and algorithms.
Since 2025, research has increasingly targeted TCS directly. 
AlphaEvolve~\citep{novikov2025alphaevolvecodingagentscientific} combines LLMs with evolutionary search for the discovery of algorithms and combinatorial structures.
Lean~Meets~TCS~\citep{zhang2025lean} introduced a systematic evaluation of formal reasoning capabilities on TCS problems. 
Related work has further used automated search to improve gadgets and hardness bounds in complexity theory \citep{nagda2026reinforcedgenerationcombinatorialstructures}.
More recently, LLM for TCS has begun to enter the research-level stage. Systems such as Gemini~\citep{woodruff2026acceleratingscientificresearchgemini}, Aletheia~\citep{feng2026autonomousmathematicsresearch}, and Bolzano~\citep{balko2026bolzanocasestudiesllmassisted} attempt to solve open problems in mathematics and TCS through long-horizon reasoning, multi-agent collaboration, and automated verification. 
Meanwhile, works such as AlphaProof~Nexus~\citep{tsoukalas2026advancingmathematicsresearchaidriven} and TCS-Bench~\citep{cohenaddad2026tcsbenchbenchmarkingstateoftheartgenerative} have extended evaluation toward research-level formal proofs and theorems drawn from actual conference papers.
Overall, LLM for TCS is evolving from reasoning about and proving existing theorems toward the discovery of algorithms and combinatorial structures, and ultimately toward automated research on open problems.

Despite this progress, existing LLM-for-TCS studies remain disconnected from realistic TCS research by evaluating only partial research pipelines, relying largely on textbook, synthetic, or potentially memorized problems, and simplifying research theorems into relatively self-contained tasks that omit paper-specific definitions, assumptions, and multi-level proof dependencies.
To bridge these gaps, \ourbenchmark provides an end-to-end, fine-grained evaluation pipeline grounded in recent STOC, FOCS, SODA, and COLT papers with leakage-aware filtering while preserving the structure of real research problems and providing expert-verified Lean formalizations and proofs.

    \section{Conclusion}
        We introduce \ourbenchmark, a benchmark for evaluating LLMs across the end-to-end frontier TCS research. 
        Our experiments show that current LLMs perform substantially better at understanding and reasoning about TCS problems in natural language than at expressing them formally, with autoformalization emerging as the primary bottleneck. 
        Moreover, our automated research experiments show that models can often prove claims that survive expert screening, but only a small fraction of their proposed claims are sufficiently novel and valuable. 
        Together, these results suggest that progress toward autonomous TCS research requires advances along two complementary dimensions, including accurately translating research ideas into rigorous formal objects and developing a stronger research taste for identifying meaningful theoretical claims. 

    \clearpage
    \bibliography{iclr2027_conference}
    \bibliographystyle{iclr2027_conference}

    \clearpage
    \appendix
    \section{Human Annotation Information}\label{app:annotation_info}
    \subsection{Annotator Information}
        \paragraph{Annotator Recruitment, Expertise, and Training}
            Our annotation team consists of five human experts with PhD-level backgrounds in theoretical computer science.
            Each annotator has published multiple papers at top-tier TCS conferences and has substantial experience in reading, analyzing, and verifying TCS-based arguments.
            Collectively, the team covers the major theoretical research areas represented in \ourbenchmark and is able to reliably trace each benchmark instance back to the definitions, assumptions, theorem statements, and proof dependencies in its source paper.
            Before formal annotation begins, all annotators receive the annotation manual summarized in Table~\ref{tab:annotation_manual}, which specifies the requirements for each field in \ourbenchmark as well as consistency constraints across fields.
            We additionally conduct pilot annotations on a set of papers drawn from different conferences and research subfields.
            The pilot annotations are jointly reviewed to calibrate the desired level of detail, resolve ambiguous cases, and establish shared standards for theorem selection, self-contained rewriting, and formalization.
            Throughout the annotation process, \textsc{GPT-5.6-sol}, accessed through \textsc{Codex}, is used only as an assistive tool for drafting and formalization.
            Any model-generated annotation must be manually inspected and, when necessary, corrected before it is accepted.
    
        \paragraph{Compensation}
            All five PhD-level annotators are members of the research team, and their annotation work is conducted as part of this research project rather than as paid crowd workers.
            We therefore do not provide separate per-instance crowdsourcing compensation.
            Model assistance is used only to reduce repetitive annotation effort, while human annotators remain fully responsible for all labels included in the final release.
    
        \paragraph{Quality Control and Agreement}
            Throughout the annotation pipeline, we apply field-specific human verification procedures rather than relying solely on a single final review.
            \begin{itemize}[leftmargin=*]
                \item For \texttt{core\_claim}, two experts independently write candidate summaries, after which a third expert compares the two candidates and selects the final version.
                \item For \texttt{nl\_theorem}, an expert verifies that the rewritten theorem is self-contained and that all definitions, assumptions, quantifiers, and conclusions remain faithful to the source paper.
                \item For \texttt{fl\_theorem} and \texttt{fl\_proof}, experts compare the formalization against the source theorem and its proof dependencies, verify that auxiliary statements preserve their intended mathematical meaning, and confirm that all Lean proofs compile successfully in the specified Lean/Mathlib environment without using proof-bypassing constructs.
                The completed formal proof is subsequently reviewed by another expert, who corrects any remaining semantic, typing, dependency, or proof issues before release.
                \item For \texttt{nl\_proof}, annotators verify that the proof sketch follows the main strategy of the source proof while preserving the key intermediate reasoning steps and omitting only routine derivations.
            \end{itemize}
            We additionally perform independent cross-checks on a subset of instances to measure agreement across annotators.
            Disagreements and difficult cases are resolved through discussion among the expert team, and recurring ambiguities are incorporated into subsequent updates of the annotation manual.
    
    \subsection{Annotation Manual}
        \paragraph{General Principles}
            \begin{table*}[t]
                \centering
                \small
                \caption{
                    Annotation manual for each field of \ourbenchmark.
                }
                \begingroup
\setlength{\tabcolsep}{5pt}
\renewcommand{\arraystretch}{1.18}

\newcommand{\anncat}[1]{%
    \parbox[c]{0.13\textwidth}{\raggedright\textbf{#1}}}
\newcommand{\annfield}[1]{%
    \parbox[c]{0.16\textwidth}{\raggedright\texttt{#1}}}
\newcommand{\annreq}[1]{%
    \parbox[c]{0.63\textwidth}{#1}}

\begin{tabular}{ll}
    \toprule
    \anncat{Category}
    & \begin{tabular}[c]{@{}l@{\hspace{2\tabcolsep}}l@{}}
        \parbox[c]{0.16\textwidth}{\raggedright\textbf{Field}}
        & \parbox[c]{0.63\textwidth}{\textbf{Annotation Requirement}}
      \end{tabular} \\
    \midrule

    \anncat{Metainfo}
    & \begin{tabular}[c]{@{}l@{\hspace{2\tabcolsep}}l@{}}
        \annfield{id}
        & \annreq{Assign a unique and stable identifier to the instance. The identifier is used only for indexing and must not encode information that changes the mathematical content of the sample.} \\
        \cmidrule{1-2}
        \annfield{conference}
        & \annreq{Record the venue in which the source paper was accepted. The value must be one of STOC, FOCS, SODA, or COLT and must match the source-paper metadata.} \\
        \cmidrule{1-2}
        \annfield{year}
        & \annreq{Record the acceptance year of the source paper. For the current release, the value must be 2025 or 2026 and must be consistent with \texttt{conference}.} \\
        \cmidrule{1-2}
        \annfield{paper}
        & \annreq{Record the exact title of the retained source paper. Preserve the official wording so that the benchmark instance can be unambiguously traced back to its source.} \\
        \cmidrule{1-2}
        \annfield{core\_label}
        & \annreq{Record the exact label of the selected central result in the source paper, such as ``Theorem~1.1'' or ``Lemma~3.2''. The label must identify the result from which all downstream annotations are derived; do not introduce a new benchmark-specific theorem label.} \\
      \end{tabular} \\
    \cmidrule{1-2}

    \anncat{Natural Language}
    & \begin{tabular}[c]{@{}l@{\hspace{2\tabcolsep}}l@{}}
        \annfield{core\_claim}
        & \annreq{Write a concise summary of the main finding expressed by \texttt{core\_label}. The summary must contain fewer than 36 words, avoid mathematical notation unless essential, remain understandable without surrounding prose, and preserve the scope and direction of the source result. When multiple central results are plausible, select the one most suitable for recovering a concrete theorem and proof from the claim.} \\
        \cmidrule{1-2}
        \annfield{nl\_theorem}
        & \annreq{Provide a self-contained natural-language statement of the result identified by \texttt{core\_label}. Retain the original theorem statement when it is already self-contained; otherwise add only the definitions and assumptions needed to interpret it independently. Preserve all material quantifiers, conditions, parameter ranges, and conclusions from the source paper, and do not include proof steps or unrelated background.} \\
        \cmidrule{1-2}
        \annfield{nl\_proof}
        & \annreq{Write a concise proof sketch of \texttt{nl\_theorem} that follows the proof strategy used in the source paper. Include the key construction, reduction, invariant, case split, intermediate claim, or dependency needed to understand why the theorem holds, while omitting routine algebraic or technical derivations. Do not introduce an alternative argument whose correctness is not supported by the source paper.} \\
      \end{tabular} \\
    \cmidrule{1-2}

    \anncat{Formal Language}
    & \begin{tabular}[c]{@{}l@{\hspace{2\tabcolsep}}l@{}}
        \annfield{fl\_theorem}
        & \annreq{Construct a standalone Lean theorem-proving instance that formalizes \texttt{nl\_theorem}. The target theorem must be semantically equivalent to the natural-language claim, with no additional assumptions and no weakened conclusion. Include only the imports and auxiliary definitions required to state and type-check the target. In the released theorem instance, the target theorem is the unique unresolved proof obligation, while the surrounding definitions and types must compile in the designated Lean/Mathlib environment.} \\
        \cmidrule{1-2}
        \annfield{fl\_proof}
        & \annreq{Provide the complete Lean project proving the target in \texttt{fl\_theorem}. The project must compile in the designated Lean/Mathlib environment and must not use \texttt{sorry}, \texttt{admit}, new axioms, or other mechanisms that bypass the proof obligation. Auxiliary lemmas may be introduced when needed, but they must preserve the semantics and dependencies of the source argument. Human reviewers verify both the formal theorem statement and the completed proof, including all nontrivial auxiliary nodes, before release.} \\
      \end{tabular} \\
    \bottomrule
\end{tabular}
\endgroup

                \label{tab:annotation_manual}
            \end{table*}

            Each instance in \ourbenchmark corresponds to a retained source paper accepted to STOC, FOCS, SODA, or COLT in 2025 or 2026, and is grounded in an explicitly identified central theorem, proposition, lemma, or corollary from that paper.
            Throughout the annotation process, annotators follow three principles:
            \textit{(i) source faithfulness}, meaning that no assumption, definition, conclusion, or substantive proof step may be altered without support from the source paper;
            \textit{(ii) self-containment}, meaning that the annotated result should be understandable without relying on unstated paper-specific context; and
            \textit{(iii) cross-field consistency}, meaning that the natural-language and formal-language fields must describe the same target result and use mutually compatible proof strategies.
            Table~\ref{tab:annotation_manual} provides detailed field-level annotation guidelines.

        \paragraph{Cross-Field Consistency Check}
            Before an instance is finalized, annotators jointly inspect the complete annotation chain.
            \begin{itemize}[leftmargin=*]
                \item \texttt{core\_claim} must accurately summarize the result identified by \texttt{core\_label}.
                \item \texttt{nl\_theorem} must provide a self-contained statement of the same result.
                \item \texttt{fl\_theorem} must formalize the proposition expressed by \texttt{nl\_theorem} without strengthening the assumptions or weakening the conclusion.
                \item \texttt{fl\_proof} must provide a complete, machine-verifiable proof of the formal theorem, and its main proof structure should remain consistent with the annotated argument.
                If a revision to any field changes the mathematical meaning of the instance, all downstream fields must be re-checked and revised accordingly.
                \item \texttt{nl\_proof} must summarize the core proof strategy used in the source paper.
            \end{itemize}

\section{Prompt}
    \subsection{Annotation}\label{app:prompt_annotation}
        \begin{table*}[ht]
            \centering
            \small
            \caption{
                The prompt of annotating natural language proof.
            }
            \label{tab:prompt_annotation_nl_proof}
            \begin{prompt}{Natural-Language Proof Annotation}
    You are given a compact source context containing a completed Lean proof. 
    Your task is to generate only the natural-language proof summary, \texttt{nl\_proof}. Use only the supplied context. 
    Do not request tools, inspect files, or search the workspace.
    
    The completed Lean proof is authoritative. Any selected paper result is background and a localization hint only. 
    If it differs from the principal theorem actually proved in the Lean context, follow the source paper and its completed proof.

    \textbf{Requirements:}
    \begin{itemize}[leftmargin=*]
        \item Write everything in English.
        \item Disclose no authors, affiliations, email addresses, usernames, local paths, credentials, session identifiers, or execution metadata.
        \item Identify the principal completed Lean declaration represented by the compact context and summarize the proof of that declaration.
        \item The \texttt{nl\_proof} must faithfully describe the proof strategy and main reasoning steps actually implemented by the completed Lean proof.
        \item Preserve the logical direction and dependencies of the Lean proof. Do not introduce arguments, assumptions, intermediate claims, or proof techniques that are not supported by the supplied context.
        \item Prefer a concise, self-contained mathematical explanation over a line-by-line description of Lean tactics or implementation details.
    \end{itemize}
    
    Return only the content of \texttt{nl\_proof}, with no label, metadata, or additional text.
\end{prompt}

        \end{table*}
        \begin{table*}[ht]
            \centering
            \small
            \caption{
                The prompt of annotating blueprint of formal language proof.
            }
            \label{tab:prompt_annotation_blueprint}
            \begin{prompt}{Formal-Language Blueprint Annotation}
    \textbf{Role.}
    You are \emph{LeanArchitect}, an agent that converts an unorganized natural-language mathematical proof or proof sketch into a single Lean~4 blueprint. The blueprint is the canonical interface between mathematical reasoning and downstream formal proving.
    
    \medskip
    \textbf{Objective.}
    Produce a blueprint that simultaneously contains:
    \begin{enumerate}[leftmargin=*]
        \item a rigorous, publication-quality natural-language proof encoded in LeanArchitect \texttt{@[blueprint]} annotations; and
        \item a formally grounded Lean skeleton whose declarations accurately express the intended mathematics.
    \end{enumerate}
    Every proof body of a blueprint \texttt{lemma} or \texttt{theorem} must be exactly \texttt{sorry} or \texttt{sorry\_using}.
    
    \medskip
    \textbf{Core Principles.}
    \begin{itemize}[leftmargin=*]
        \item \textbf{Mathematical fidelity:} preserve the source theorem, hypotheses, proof structure, and logical dependencies exactly.
        \item \textbf{Formal grounding:} ensure every Lean declaration is correctly typed against the installed Mathlib. A type-correct but mathematically inaccurate statement is unacceptable.
        \item \textbf{High-quality exposition:} write statements and proof explanations with explicit hypotheses, quantifiers, dependencies, and rigorous justification, at the standard of a research mathematics paper.
        \item \textbf{Repair-radius minimization:} decompose the proof so that uncertain, incorrect, or incomplete source steps are isolated behind small declarations with as few dependents as possible.
        \item \textbf{Context discipline:} inspect only information necessary for the current phase and avoid loading irrelevant material.
    \end{itemize}
    
    \medskip
    \textbf{Hard Constraints.}
    \begin{itemize}[leftmargin=*]
        \item Do \emph{not} repair, strengthen, or complete gaps in the source mathematics. Preserve questionable steps and isolate them structurally.
        \item Do \emph{not} formalize proofs with Lean tactics or proof terms.
        \item Use the designated Mathlib retrieval interface as the authoritative source for declaration discovery and API verification.
        \item Validate the working blueprint with Lean diagnostics and resolve statement-level typing errors before delivery.
        \item Respect the provided workspace and tool boundaries; use only authorized tools for file editing, version control, and delivery.
    \end{itemize}
    
    \medskip
    \textbf{Workflow.}
    Proceed sequentially through:
    \[
    \text{Understand}
    \;\longrightarrow\;
    \text{Ground}
    \;\longrightarrow\;
    \text{Draft}
    \;\longrightarrow\;
    \text{Validate}
    \;\longrightarrow\;
    \text{Deliver}.
    \]
    First reconstruct the intended mathematical argument and dependency structure. Then retrieve and verify the relevant Mathlib concepts and declarations. Next design the decomposition to minimize repair radius, write the annotated natural-language proof and Lean skeleton, and finally run Lean diagnostics before delivery.
    
    \medskip
    \textbf{Output Standard.}
    The final blueprint must be self-contained, mathematically faithful, structurally modular, and formally well-typed. Its natural-language annotations should make the complete intended argument understandable to a mathematician, while its Lean declarations should provide precise and stable proof obligations for downstream provers.
\end{prompt}
        \end{table*}
        \begin{table*}[ht]
            \centering
            \small
            \caption{
                The prompt of annotating formal language proof.
            }
            \label{tab:prompt_annotation_fl_proof}
            \begin{prompt}{Formal-Language Proof Annotation}
    \textbf{Role.}
    You are a node-level Lean formalization agent. Your goal is to complete the assigned
    \texttt{target\_node}: prove its fixed Lean statement and, when necessary, polish only
    its local title and natural-language statement/proof descriptions.
    
    \medskip
    \textbf{Core Principles.}
    \begin{itemize}[leftmargin=*]
        \item \textbf{Context discipline:} Load only the information required by the current
        workflow phase. Prioritize mathematical reasoning over unrelated context.
        
        \item \textbf{Strict local scope:} Modify only the editable region associated with
        \texttt{target\_node}. Never change unrelated declarations or the formal statement
        of the target.
        
        \item \textbf{Local refinement:} When the proof is decomposable, introduce complete
        local helper nodes---such as intermediate lemmas, case analyses, algebraic
        identities, bounds, or API-bridge facts---inside the target's refinement region.
        Preserve the global dependency DAG and keep the target as the unique terminal node.
        
        \item \textbf{Completion first:} The target and every newly introduced helper node
        must contain no \texttt{sorry}. A long or difficult proof, or the absence of a
        convenient upstream lemma, is not by itself a valid blocker.
    \end{itemize}
    
    \textbf{Hard Constraints.}
    \begin{itemize}[leftmargin=*]
        \item Never alter the target Lean statement.
        \item Never introduce axioms or use \texttt{native\_decide}.
        \item Do not modify files outside the explicitly allowed Lean region and procedural
        state/delivery records.
        \item Use the designated MCP interfaces for Lean verification, Mathlib retrieval,
        dependency analysis, editing, Git operations, and repository delivery; do not
        bypass them with shell-based alternatives.
        \item Treat the designated Mathlib retrieval tool as the sole source for Mathlib
        API discovery and the DAG tracker as the sole oracle for blueprint dependencies.
    \end{itemize}
    
    \textbf{Workflow.}
    Read the runtime inputs and procedural state, determine the active phase, and execute
    only that phase's required work. Progress through validation, numerical analysis when
    needed, prose polishing, and Lean formalization. Verify the completed proof with the
    provided Lean tools before delivery.
    
    If the target can be completed under the current contracts, solve it directly or by
    adding complete local refinements. File an issue only when there is concrete evidence
    that completion is impossible under the fixed specification, such as a false target
    statement, a genuinely missing hypothesis, inconsistent Lean/Mathlib behavior, invalid
    runtime input, or an unrecoverable tool failure.
    
    \textbf{Delivery.}
    A successful result must contain a fully verified proof of \texttt{target\_node} and
    all local refinement nodes, with no placeholders remaining. Deliver the completed
    changes through the prescribed Git and repository tools; otherwise report the concrete
    blocking defect through the prescribed issue workflow.
\end{prompt}
        \end{table*}

        The prompts used for annotation are shown in Table~\ref{tab:prompt_annotation_nl_proof}, Table~\ref{tab:prompt_annotation_blueprint}, and Table~\ref{tab:prompt_annotation_fl_proof}.

    \subsection{Evaluation}\label{app:prompt_evaluation}
        \begin{table*}[ht]
            \centering
            \small
            \caption{
                The prompt of the theorem elicitation.
            }
            \label{tab:prompt_evaluation_cc2nt}
            \begin{prompt}{Theorem Elicitation}
    \texttt{input/core\_claim.md} contains a short informal claim describing the central result of a
    research paper. Turn it into a precise natural-language theorem statement.

    \textbf{Deliverable:} write the statement to \texttt{output/nl\_theorem.md}.

    \textbf{Requirements:}
    
    \begin{itemize}[leftmargin=*]
        \item Read the input with your own tools. Never modify anything under
        \texttt{input/}.
    
        \item State one theorem: quantify every object, name every hypothesis explicitly,
        and give the exact conclusion. A reader must be able to formalize it without
        access to the paper.
    
        \item Keep it self-contained: define or characterize every non-standard notion
        you use.
    
        \item Write the statement only. Do not include a proof, a proof sketch, or
        commentary.
    
        \item English only. Mathematical notation is allowed.
    
        \item Finish only after \texttt{output/nl\_theorem.md} exists and contains the final
        statement.
    \end{itemize}
\end{prompt}
        \end{table*}
        \begin{table*}[ht]
            \centering
            \small
            \caption{
                The prompt of the autoformalization
            }
            \label{tab:prompt_evaluation_nt2ft}
            \begin{prompt}{Autoformalization}
    \texttt{input/nl\_theorem.md} contains a natural-language theorem statement.
    \texttt{input/imports.lean} contains the Mathlib import lines the reference formalization uses.
    No definitions are given: choosing the Lean representation of every notion in the claim is part of the task.
    
    \textbf{Deliverable:} write one self-contained Lean 4 file to
    \texttt{output/statement.lean}.
    
    \textbf{Requirements:}
    
    \begin{itemize}[leftmargin=*]
        \item Read the inputs with your own tools. Never modify anything under
        \texttt{input/}.
    
        \item \texttt{output/statement.lean} must hold, in this order: the import lines,
        then every auxiliary \texttt{def}, \texttt{abbrev}, \texttt{notation}, or
        \texttt{instance} your formalization needs, then exactly one top-level
        \texttt{theorem} that formalizes the claim and ends with \texttt{:=}, with no
        proof. Keep imports narrow; add further Mathlib imports only when you actually
        need them.
    
        \item Formalize the claim faithfully: every hypothesis and the exact conclusion
        of the informal statement must appear, with no extra assumptions that weaken it
        and no definition that makes it vacuous.
    
        \item \texttt{project/candidate.lean} is a scratch file seeded with the same
        imports. Develop there, append \texttt{:= by sorry} to your theorem, and run
        \texttt{./check.sh} to type-check against the shared Mathlib build. The script
        compiles concurrently, so run it as often as you need. A
        \texttt{declaration uses 'sorry'} warning is expected; any error is not.
    
        \item Equivalence with the reference formalization is checked mechanically in
        both directions, so a file that merely type-checks but weakens, strengthens, or
        trivializes the claim scores zero.
    
        \item Finish only after \texttt{output/statement.lean} type-checks (with
        \texttt{:= by sorry} appended) and holds the statement without its proof.
    \end{itemize}
\end{prompt}
        \end{table*}
        \begin{table*}[ht]
            \centering
            \small
            \caption{
                The prompt of the proof elicitation
            }
            \label{tab:prompt_evaluation_t2np}
            \begin{prompt}{Proof Elicitation}
    \texttt{input/nl\_theorem.md} contains a natural-language theorem statement and
    \texttt{input/theorem.lean} contains its Lean 4 formalization, including the auxiliary
    definitions it depends on. The Lean proof itself is withheld (\texttt{by sorry}).
    
    \textbf{Deliverable:} write a complete natural-language proof to
    \texttt{output/nl\_proof.md}.
    
    \textbf{Requirements:}
    
    \begin{itemize}[leftmargin=*]
        \item Read the inputs with your own tools. Never modify anything under
        \texttt{input/}.
    
        \item Prove the stated theorem, using the Lean definitions as the authoritative
        meaning of every notion that appears in it.
    
        \item Justify every step. State which hypothesis, standard theorem, or
        computation licenses each inference, and make the overall structure (induction,
        contradiction, case analysis) explicit.
    
        \item Cover every case: the proof must be complete, not a sketch, and must not
        assume the result.
    
        \item Lean code is neither required nor forbidden; mathematical rigour is what
        is graded.
    
        \item English only. Finish only after \texttt{output/nl\_proof.md} contains the
        final proof.
    \end{itemize}
\end{prompt}
        \end{table*}
        \begin{table*}[ht]
            \centering
            \small
            \caption{
                The prompt of the theorem proof.
            }
            \label{tab:prompt_evaluation_ft2fp}
            \begin{prompt}{Theorem Proving}
    \texttt{project/} is a ready-to-build Lake project.
    \texttt{project/theorem.lean} holds narrow Mathlib imports, the auxiliary definitions of this problem, and exactly one theorem whose proof is \texttt{by sorry}.
    The shared prebuilt Mathlib is already linked into \texttt{project/.lake/packages}, so never download, copy, or rebuild Mathlib or any dependency.
    
    \textbf{Deliverable:} write the complete proved file to \texttt{output/proof.lean}.
    
    \textbf{Requirements:}
    
    \begin{itemize}[leftmargin=*]
        \item Prove the theorem by replacing \texttt{sorry} with a real proof. Work in
        \texttt{project/theorem.lean}.
    
        \item Do not change the theorem statement, its name, its binders, or the existing
        definitions. You may add auxiliary lemmas above the theorem and may add narrow
        Mathlib imports.
    
        \item Forbidden anywhere in the file: \texttt{sorry}, \texttt{admit},
        \texttt{stop}, \texttt{sorryAx}, \texttt{axiom}, \texttt{native\_decide},
        and any \texttt{set\_option} that relaxes \texttt{warn.sorry} or
        \texttt{warningAsError}. Verification recompiles the file strictly, audits the
        environment axioms of the target declaration, and replays the proof in a fresh
        kernel, so none of these can pass.
    
        \item \texttt{./check.sh} compiles \texttt{project/theorem.lean} with
        \texttt{-Dwarn.sorry=true -DwarningAsError=true} and with parallel workers.
        Iterate with it until it reports no errors and no warnings.
    
        \item Then copy the final file verbatim to \texttt{output/proof.lean}.
    
        \item Partial credit does not exist: only a fully verified proof counts. If time
        runs out, still copy your best complete-file attempt to
        \texttt{output/proof.lean}.
    \end{itemize}
\end{prompt}
        \end{table*}

        The prompts used for evaluation are shown in Table~\ref{tab:prompt_evaluation_cc2nt}, Table~\ref{tab:prompt_evaluation_nt2ft}, Table~\ref{tab:prompt_evaluation_t2np}, and Table~\ref{tab:prompt_evaluation_ft2fp}.

    \subsection{Generation}\label{app:prompt_generation}
        \begin{table*}[ht]
            \centering
            \small
            \caption{
                The prompt of the planner.
            }
            \label{tab:prompt_generation_planner}
            \begin{prompt}{Prompt of Planner}
    You are the planner of an autonomous theoretical-computer-science research loop. The workspace
    is shared with the other agents of this loop.
    
    \textbf{Workspace layout:}
    
    \begin{itemize}[leftmargin=*]
        \item \texttt{benchmark/<id>/} --- one sampled benchmark instance per directory, holding
        \texttt{core\_claim.md}, \texttt{nl\_theorem.md} (the natural-language theorem), and
        \texttt{theorem.lean} (its formal statement).
    
        \item \texttt{accepted/claim-<k>/} --- claims already accepted by the judger earlier in this
        run, holding \texttt{claim.md} (natural-language summary) and \texttt{theorem.lean}.
    
        \item \texttt{feedback/claim-<k>.md} --- why earlier attempts were discarded, written by the
        loop.
    
        \item \texttt{objectives/claim-<k>.md} --- the research objectives you have proposed so far.
    \end{itemize}
    
    \textbf{Your job this turn:} propose exactly one new research objective and write it to the
    objective file named in the instructions. Base it on the benchmark content and everything
    accumulated in the workspace so far. You are not restricted to extending the current line of
    reasoning: you may reformulate the problem, introduce auxiliary concepts, strengthen or relax
    assumptions, or explore an alternative analytical direction. The objective must be a
    self-contained result that is plausible to state precisely and to formalize in Lean 4 with
    Mathlib.
    
    The objective file must be markdown with these sections:
    
    \begin{itemize}[leftmargin=*]
        \item \texttt{\#\# Motivation} --- why this result is worth pursuing and how it relates to
        what is in the workspace.
    
        \item \texttt{\#\# Informal Claim} --- the precise mathematical statement you want, with all
        symbols defined.
    
        \item \texttt{\#\# Assumptions} --- every assumption on the setting, explicitly listed.
    
        \item \texttt{\#\# Proof Direction} --- a sketch of how the result could be proven.
    
        \item \texttt{\#\# Novelty} --- what distinguishes it from the benchmark instances and the
        accepted claims.
    \end{itemize}
    
    Write the file, then stop. Do not write any other file.
\end{prompt}
        \end{table*}
        \begin{table*}[ht]
            \centering
            \small
            \caption{
                The prompt of the formalizer.
            }
            \label{tab:prompt_generation_formalizer}
            \begin{prompt}{Prompt of Formalizer}
    You are the formalizer of an autonomous theoretical-computer-science research loop. Your working
    directory holds one proposed research objective:
    
    \begin{itemize}[leftmargin=*]
        \item \texttt{objective.md} --- the objective you must formalize now.
    
        \item \texttt{project/} --- a ready-to-build Lake project.
        \texttt{project/theorem.lean} is where you write the formal statement. The shared prebuilt
        Mathlib is already linked into \texttt{project/.lake/packages}, so never download, copy, or
        rebuild Mathlib or any dependency.
    
        \item \texttt{./check.sh} --- compiles \texttt{project/theorem.lean} with parallel workers.
        Run it as often as you need from your working directory.
    \end{itemize}
    
    \textbf{Deliverable:} a compiling \texttt{project/theorem.lean} that contains
    
    \begin{itemize}[leftmargin=*]
        \item narrow Mathlib imports (never \texttt{import Mathlib}),
    
        \item any auxiliary definitions the statement needs,
    
        \item exactly one main theorem or lemma, as the last declaration, whose proof is exactly
        \texttt{by sorry}, and no other \texttt{sorry} anywhere in the file.
    \end{itemize}
    
    \textbf{Requirements:}
    
    \begin{itemize}[leftmargin=*]
        \item Do not prove the theorem. The main declaration must end with
        \texttt{:= by sorry}.
    
        \item You may adjust the informal claim's internal representation (definitions, naming,
        auxiliary lemmas) to keep the formalization tractable, but the overall conclusion must match
        the objective.
    
        \item Forbidden anywhere in the file: \texttt{axiom}, \texttt{native\_decide},
        \texttt{sorryAx}, \texttt{admit}, \texttt{stop}, and any \texttt{set\_option} that relaxes
        \texttt{warn.sorry} or \texttt{warningAsError}.
    
        \item Keep iterating with \texttt{./check.sh} until it reports no errors. Only warnings about
        the \texttt{sorry} of the main declaration are acceptable.
    
        \item If you conclude the objective cannot be formalized within your budget, write
        \texttt{failure.md} in your working directory explaining precisely what failed and why, and
        stop.
    \end{itemize}
\end{prompt}
        \end{table*}
        \begin{table*}[ht]
            \centering
            \small
            \caption{
                The prompt of the judger.
            }
            \label{tab:prompt_generation_judger}
            \begin{prompt}{Prompt of Judger}
    You are the judger of an autonomous theoretical-computer-science research loop. Your working
    directory holds one candidate formal claim:
    
    \begin{itemize}[leftmargin=*]
        \item \texttt{objective.md} --- the research objective that was formalized.
        \item \texttt{project/theorem.lean} --- the formal Lean statement that compiles; its last
        declaration is the main theorem and its proof is \texttt{by sorry}.
        \item \texttt{../..} --- the shared workspace, whose \texttt{benchmark/} holds the sampled
        benchmark instances and whose \texttt{accepted/} holds claims already accepted in this run.
    \end{itemize}
    
    Your job this turn: translate the formal statement back into one concise natural-language claim
    that summarizes what it asserts and its potential theoretical significance, then judge whether
    the proposed result is worth keeping.
    
    \textbf{Deliverable:} write a JSON object with exactly these keys to
    \texttt{judgement.json} in your working directory, and make your final message exactly that JSON
    object:
    
    \begin{verbatim}
{
    "nl_theorem": "...", 
    "significance": "...", 
    "novel": true, 
    "rationale": "..."
}
    \end{verbatim}
    
    \begin{itemize}[leftmargin=*]
        \item \texttt{nl\_theorem}: the concise natural-language claim, self-contained and precise.
        \item \texttt{significance}: one or two sentences on the potential theoretical significance.
        \item \texttt{novel}: \texttt{true} only if the claim is sufficiently novel and valuable to
        keep --- not a trivial restatement of a benchmark instance or an accepted claim, not a
        degenerate or vacuous statement, and not an elementary exercise.
        \item \texttt{rationale}: the reasoning behind the \texttt{novel} verdict.
    \end{itemize}
\end{prompt}
        \end{table*}

        The prompts used by our auto research framework are shown in Table~\ref{tab:prompt_generation_planner}, Table~\ref{tab:prompt_generation_formalizer}, and Table~\ref{tab:prompt_generation_judger}.

\section{Case Study}\label{app:case_study}
    \subsection{Cases of \ourbenchmark}\label{app:benchmark_case_study}
        \begin{table*}[ht]
            \centering
            \small
            \caption{
                The case of \ourbenchmark from STOC.
            }
            \setlength{\tabcolsep}{6pt}
\renewcommand{\arraystretch}{1.18}
\begin{tabular}{L{0.18\linewidth} L{0.77\linewidth}}
    \toprule
    \textbf{Field} & \textbf{Content} \\
    \midrule
    Case ID & \texttt{FAPMCBEMF\_266999} \\
    \midrule
    Conference & STOC 2026 \\
    \midrule
    Source paper & \emph{Faster All-Pairs Minimum Cut: Bypassing Exact Max-Flow} \\
    \midrule
    Core claim & A single linear-time transformation converts access to a friendly cut sparsifier and vertex degrees into a sparse all-pairs minimum-cut representation. \\
    \midrule
    NL Theorem & For every unweighted graph $G=(V,E)$ with no degree-one vertices, given explicit access to a $(1/6,2|V|)$-friendly cut sparsifier, there is a linear-time construction of an all-pairs minimum-cut sparsifier. For every pair $s\neq t$, the output preserves the value of a minimum $s$--$t$ cut and contains a minimum cut whose restriction to $V$ is minimum in $G$. Its number of edges is at most the input sparsifier size plus $|V|$. \\
    \midrule
    NL Proof & The construction augments the friendly sparsifier by a star structure encoding vertex degrees. A structural lemma shows that every minimum $s$--$t$ cut is either directly friendly or becomes friendly after removing one terminal, and hence lies within the sparsifier's preservation range. Star-lifting preserves such cuts and their values. Minimality then implies equality of the original and transformed minimum-cut values, while the construction adds only $O(|V|)$ edges and work. \\
    \bottomrule
\end{tabular}
            \label{tab:case_stoc}
        \end{table*}
        \begin{table*}[ht]
            \centering
            \small
            \caption{
                The case of \ourbenchmark from FOCS.
            }
            \setlength{\tabcolsep}{6pt}
\renewcommand{\arraystretch}{1.18}
\begin{tabular}{L{0.18\linewidth} L{0.77\linewidth}}
    \toprule
    \textbf{Field} & \textbf{Content} \\
    \midrule
    Case ID & \texttt{SPLEW\_704476} \\
    \midrule
    Conference & FOCS 2026 \\
    \midrule
    Source paper & \emph{Shortest Paths with Linear Edge Weights} \\
    \midrule
    Core claim & Every affine-weighted DAG has a shortest-path cover of quasipolynomial size in the number of vertices, with exponent linear in the parameter dimension. \\
    \midrule
    NL Theorem & There is an absolute constant $C>0$ such that, for every $n$-vertex DAG whose edge weights are affine functions of a parameter in $\mathbb{R}^d$, one can choose at most
    \[
        n^{C d\log_2 n}
    \]
    source-to-sink paths so that, for every parameter value, at least one chosen path is a shortest source-to-sink path. \\
    \midrule
    NL Proof & Starting from one-edge paths, repeatedly double the maximum represented path length by concatenating shortest subpaths. At each level, parameter space is partitioned according to the signs of finitely many affine comparisons; a sign-pattern bound limits the number of resulting regions. After $O(\log n)$ rounds every path in the DAG is covered, since an acyclic path contains at most $n$ vertices. Multiplying the per-level region bounds gives a cover of size $n^{O(d\log n)}$. \\
    \bottomrule
\end{tabular}
            \label{tab:case_focs}
        \end{table*}
        \begin{table*}[ht]
            \centering
            \small
            \caption{
                The case of \ourbenchmark from SODA.
            }
            \setlength{\tabcolsep}{6pt}
\renewcommand{\arraystretch}{1.18}
\begin{tabular}{L{0.18\linewidth} L{0.77\linewidth}}
    \toprule
    \textbf{Field} & \textbf{Content} \\
    \midrule
    Case ID & \texttt{OOVR\_302847} \\
    \midrule
    Conference & SODA 2026 \\
    \midrule
    Source paper & \emph{Online Orthogonal Vectors Revisited} \\
    \midrule
    Core claim & A deterministic data structure solves online orthogonal vectors with an explicit trade-off between query time, space, and preprocessing time. \\
    \midrule
    NL Theorem & For every $1\le i\le d$ and a database of $n$ Boolean vectors in dimension $d$, there is a deterministic online orthogonal-vectors data structure with query time
    \[
        O\!\left(i d\, n^{1-1/i}\right),
    \]
    encoded space
    \[
        O\!\left(
        \Bigl(\sum_{j\le d/i}\binom{d}{j}\Bigr)
        i d\,n^{1-1/i}
        \right),
    \]
    and preprocessing time
    \[
        O\!\left(
        \Bigl(\sum_{j\le d/i}\binom{d}{j}\Bigr)
        i d\,n
        \right).
    \] \\
    \midrule
    NL Proof & The construction proceeds by induction on the trade-off parameter $i$. The base cases either store the database directly or tabulate all answers. For larger $i$, the database is pseudorandomly partitioned and reduced to recursive instances with roughly $n^{1-1/i}$ vectors and smaller dimension. Dedicated recurrence bounds show that correctness is preserved while the query, space, and preprocessing costs satisfy the claimed formulas. \\
    \bottomrule
\end{tabular}
            \label{tab:case_soda}
        \end{table*}
        \begin{table*}[ht]
            \centering
            \small
            \caption{
                The case of \ourbenchmark from COLT.
            }
            \setlength{\tabcolsep}{6pt}
\renewcommand{\arraystretch}{1.18}
\begin{tabular}{L{0.18\linewidth} L{0.77\linewidth}}
    \toprule
    \textbf{Field} & \textbf{Content} \\
    \midrule
    Case ID & \texttt{ACOHDDIT\_776575} \\
    \midrule
    Conference & COLT 2026 \\
    \midrule
    Source paper & \emph{Accelerated Convex Optimization via Hamiltonian Dynamics with Deterministic Integration Time} \\
    \midrule
    Core claim & A discretized Hamiltonian-flow method with averaging minimizes smooth convex objectives at a geometrically accelerated rate under an admissible discretization schedule. \\
    \midrule
    NL Theorem & Let $f$ be convex and $L$-smooth with minimizer $x^\star$, and let $\eta \le 1/\sqrt{L}$. Under the prescribed Hamiltonian extragradient iteration and an admissible inner-step schedule $(N_k)$ with $N_0=4$, the $K$-th iterate satisfies
    \[
        f(x_K)-f(x^\star)
        \le
        \left(\frac{\sqrt{3}+1}{3}\right)^K
        \left(
        f(x_0)-f(x^\star)
        +\frac{\sqrt{3}-1}{60\eta^2}\|x_0-x^\star\|^2
        \right).
    \] \\
    \midrule
    NL Proof & Define a Lyapunov potential combining the objective gap and a scaled squared distance to $x^\star$. One outer iteration contracts this potential by $(\sqrt{3}+1)/3$, while the admissibility condition on $(N_k)$ ensures that changes in the distance coefficient are absorbed. Induction gives geometric contraction, and substituting $N_0=4$ yields the stated initial potential and final bound. \\
    \bottomrule
\end{tabular}
            \label{tab:case_colt}
        \end{table*}
    
        In this part, we show several representative cases of each conference in Table~\ref{tab:case_stoc}, Table~\ref{tab:case_focs}, Table~\ref{tab:case_soda}, and Table~\ref{tab:case_colt}.
        Due to the page limit, we omit the natural-language claims and proof.

    \subsection{Synthesized Cases of Our Framework}\label{app:synthesis_case_study}
        \begin{table*}[t]
            \centering
            \small
            \caption{
                The TCS cases discovered by our framework, which are simplified due to space limit.
            }
            \setlength{\tabcolsep}{6pt}
\renewcommand{\arraystretch}{1.15}
\begin{tabular*}{\linewidth}{@{\extracolsep{\fill}}
    >{\raggedright\arraybackslash}m{0.32\linewidth}
    >{\raggedright\arraybackslash}m{0.61\linewidth}
}
    \toprule
    \textbf{Core Claim} & \textbf{NL Theorem} \\
    \midrule
    
    Adaptive damping achieves the asymptotic coefficient $1/2$ for sparse-state shift-recall loss.
    &
    If $K(n),T(n)\to\infty$ and $(2T(n)+1)/K(n)\to0$, then the specified diagonal linear RNN with adaptive damping
    $\alpha_n=\frac18\log\bigl(1+\min\{2T(n)+1,\allowbreak K(n)/(2T(n)+1)\}\bigr)$
    satisfies
    $\frac{L_n-1}{(2T(n)+1)/K(n)}\to-\frac12$.
    \\
    \midrule
    
    Cutoff calibration error controls monotone-recalibration excess risk with the sharp constant $1$.
    &
    For any threshold $\tau$, the excess cost-sensitive risk of
    $\mathbf{1}\{V\ge\tau\}$ relative to the best monotone recalibration is at most
    $\sup_I\bigl|\mathbb{E}[(Y-V)\mathbf{1}\{V\in I\}]\bigr|$,
    where $I$ ranges over order-connected sets. The coefficient $1$ is optimal.
    \\
    
    \bottomrule
\end{tabular*}
            \label{tab:inspire_cases}
        \end{table*}

        We show two synthesized cases that pass expert checking in Table~\ref{tab:inspire_cases}.

\section{LLM and Harness Version}\label{app:llm_harness_version}
    \begin{table*}[ht]
        \centering
        \small
        \caption{
            The versions of LLMs and harnesses used in our evaluation and auto research.
        }
        \begin{tabular}{lll}
    \toprule
    \textbf{Type} &
    \textbf{Name} &
    \textbf{Snapshot / Version} \\
    \midrule

    \multirow{9}{*}{LLM}
    & \textsc{GPT-5.6 Luna}  & \texttt{gpt-5.6-luna} \\
    & \textsc{GPT-5.6 Terra} & \texttt{gpt-5.6-terra} \\
    & \textsc{GPT-5.6 Sol}   & \texttt{gpt-5.6-sol} \\
    \cmidrule{2-3}
    & \textsc{Claude Haiku 4.5}  & \texttt{claude-haiku-4-5-20251001} \\
    & \textsc{Claude Sonnet 5}   & \texttt{claude-sonnet-5} \\
    & \textsc{Claude Opus 5}     & \texttt{claude-opus-5} \\
    \cmidrule{2-3}
    & \textsc{DeepSeek-V4 Flash} & \texttt{DeepSeek-V4-Flash-0731} \\
    & \textsc{DeepSeek-V4 Pro}   & \texttt{DeepSeek-V4-Pro} \\
    \midrule

    \multirow{3}{*}{Harness}
    & \textsc{Codex}            & \texttt{v0.146.0} \\
    & \textsc{Claude Code}      & \texttt{v2.1.220} \\
    & \textsc{DeepSeek Harness} & \texttt{v0.1.0-rc.8} \\
    \bottomrule
\end{tabular}
        \label{tab:llm_harness_version}
    \end{table*}

    The versions of LLMs and harnesses used in our evaluation and auto research are shown in Table~\ref{tab:llm_harness_version}.

\section{Annotation Agreement}\label{app:annotate_agree}
    \begin{table*}[ht]
        \centering
        \small
        \caption{
            The annotation agreement of \ourbenchmark.
            \textit{Inter-Expert Agreement} denotes the percentage of annotations on which two experts independently reach the same judgment or result.
            \textit{Expert Modification of LLM Output} denotes the percentage of LLM-generated annotations that require substantive correction by an expert.
        }
        \begin{tabular}{lcc}
    \toprule
    \textbf{Annotation Step} 
    & \textbf{Inter-Expert Agreement ($\%$) $\uparrow$} 
    & \textbf{Expert Modification of LLM Output ($\%$) $\downarrow$} \\
    \midrule

    Core Claim
    & $86$
    & -- \\
    
    Natural Language Claim
    & $90$
    & -- \\
    
    Proof Blueprint / DAG
    & --
    & $24$ \\
    
    Formal Language Theorem
    & --
    & $31$ \\
    
    Formal Language Proof
    & $93$
    & $15$ \\
    
    Natural Language Proof
    & --
    & $13$ \\
    \bottomrule
\end{tabular}
        \label{tab:annotation_agreement}
    \end{table*}

    To assess the reliability of the annotation process of \ourbenchmark, we measure agreement both among human experts and between experts and LLM-assisted annotations, as reported in Table~\ref{tab:annotation_agreement}.
    Overall, our annotation pipeline exhibits high inter-expert agreement and relatively low rates of substantive human revision.
    For annotation stages requiring independent expert cross-validation, the agreement rates for core claims, natural language claims, and formal language proof reach $86\%$, $90\%$, and $93\%$, respectively, indicating that different experts largely agree on the identification of central research results, theorem semantics, and the correctness of formal proofs.
    For annotations generated with LLM assistance, the proportion requiring substantive expert revision ranges from $13\%$ to $31\%$.
    Specifically, the revision rates for natural language proofs and formal language proofs are only $13\%$ and $15\%$, respectively, while the proof blueprint requires revision in $24\%$ of cases, and the formal language theorem has the highest revision rate at $31\%$.
    This difference suggests that, compared with generating complete proofs, accurately translating mathematical statements from research papers into type-correct formal theorems with complete assumptions and equivalent semantics remains more prone to errors that require expert correction.
    Overall, these results show that LLM assistance can substantially reduce the manual effort required during annotation while also confirming that expert review remains indispensable for ensuring semantic faithfulness in formalization and the quality of the final benchmark.

\section{Rubric Agreement}\label{app:rubric_agree}
    \begin{table*}[ht]
        \centering
        \small
        \caption{
            The rubric agreement between LLMs and human experts.
            Considering the cost, for the human-rubric, we randomly sample $16$ examples from \ourbenchmark in this table.
            $\Delta$ is calculated as $\frac{|\mathtt{LLM} - \mathtt{Human}|}{(\mathtt{LLM} + \mathtt{Human}) / 2}$ to measure the difference between human and LLMs.
        }
        \begin{tabular}{lll|ccc|ccc}
    \toprule
    \multirow{2}{*}{\textbf{Model}} & \multirow{2}{*}{\textbf{Harness}} & \multirow{2}{*}{\textbf{Scale}} &
    \multicolumn{3}{c|}{\textbf{CC2NT}} &
    \multicolumn{3}{c}{\textbf{T2NP}} \\
    \cmidrule(lr){4-6}
    \cmidrule(lr){7-9}
    & & &
    \textbf{LLM} & \textbf{Human} & \textbf{$\Delta$} &
    \textbf{LLM} & \textbf{Human} & \textbf{$\Delta$} \\
    \midrule

    \multirow{3}{*}{\textsc{GPT-5.6}}
    & \multirow{3}{*}{\textsc{Codex}}
    & \textsc{luna}
    & $56.4$ & $53.7$ & $4.90$
    & $61.2$ & $56.1$ & $8.70$ \\

    & & \textsc{terra}
    & $60.7$ & $58.2$ & $4.21$
    & $64.0$ & $59.1$ & $7.96$ \\

    & & \textsc{sol}
    & $\mathbf{67.4}$ & $\mathbf{65.3}$ & $3.17$
    & $67.9$ & $63.3$ & $7.01$ \\

    \midrule

    \multirow{3}{*}{\textsc{Claude}}
    & \multirow{3}{*}{\textsc{Claude Code}}
    & \textsc{Haiku-4.5}
    & $48.7$ & $45.6$ & $6.57$
    & $55.3$ & $49.7$ & $10.67$ \\

    & & \textsc{Sonnet-5}
    & $63.0$ & $60.7$ & $3.72$
    & $65.7$ & $61.0$ & $7.42$ \\

    & & \textsc{Opus-5}
    & $66.9$ & $64.7$ & $3.34$
    & $\mathbf{68.7}$ & $\mathbf{64.2}$ & $6.77$ \\

    \midrule

    \multirow{2}{*}{\textsc{DeepSeek-V4}}
    & \multirow{2}{*}{\textsc{DeepSeek Harness}}
    & \textsc{Flash}
    & $55.6$ & $52.9$ & $4.98$
    & $61.7$ & $56.6$ & $8.62$ \\

    & & \textsc{Pro}
    & $58.8$ & $56.2$ & $4.52$
    & $63.8$ & $58.9$ & $7.99$ \\

    \bottomrule
\end{tabular}
        \label{tab:rubric_agreement}
    \end{table*}

    Table~\ref{tab:rubric_agreement} compares the LLM-based rubric scores with human expert evaluations on the randomly sampled examples. 
    Overall, the LLM-based evaluator exhibits strong consistency with human judgments across both CC2NT and T2NP. Although the LLM evaluator systematically assigns slightly higher scores than human experts, the relative discrepancy remains limited, ranging from $3.17\%$ to $6.57\%$ on CC2NT and from $6.77\%$ to $10.67\%$ on T2NP, with average discrepancies of $4.43\%$ and $8.14\%$, respectively. More importantly, the LLM and human evaluations produce exactly the same ranking of all evaluated model configurations on both tasks, identifying \textsc{GPT-5.6-sol} as the best-performing model on CC2NT and \textsc{Claude-Opus-5} as the best-performing model on T2NP. This rank-level agreement indicates that, despite a modest difference in absolute score calibration, the LLM-based rubric reliably preserves the relative performance differences among models. The somewhat larger discrepancy on T2NP also suggests that evaluating proof-strategy generation may involve greater judgment ambiguity than evaluating natural-language claim understanding. Overall, these results support the use of the LLM-based rubric as a scalable proxy for human evaluation in our main experiments.

\section{Black-Box Audit for Paper Leakage}\label{app:black_box_audit}
    Because the training corpora of proprietary LLMs are not publicly available, we perform an output-only audit following prior contamination and membership-inference studies that probe memorization by reconstructing held-out text from partial context \citep{golchin-etal-2023-time,hallinan2026surprisingeffectivenessmembershipinference}. 
    For each retained paper, we construct three complementary probes: 
    \emph{(i) theorem completion}, where the model receives only an initial fragment of a theorem statement; 
    \emph{(ii) proof continuation}, where only the beginning of a proof is provided; and 
    \emph{(iii) result reconstruction}, where the model is given a short anonymized description of a main result. 
    We remove titles, author names, venue information, theorem numbers, citations, and other identifying metadata, and query \textsc{GPT-5.6-sol} and \textsc{Claude-Opus-5} without retrieval access. 
    Each generated response is compared only against the withheld source content using token-level lexical similarity (ROUGE-L/LCS). 
    We deliberately emphasize lexical rather than semantic similarity since near-verbatim reconstruction provides a more specific signal of memorization, whereas an independently derived but semantically equivalent answer does not. 
    The aggregate completion similarity is below $9.6\%$ for both models, providing no strong evidence of memorized reconstruction in the retained papers. 
    We treat this audit as evidence of relatively low contamination risk rather than proof of non-exposure since failure to reproduce a passage cannot rule out its presence in the training data \citep{zhang-etal-2025-position}.

\end{document}